\documentclass{article} 
\usepackage{arxiv}

\usepackage{amsmath,amsfonts,bm}

\def\eqref#1{equation~\ref{#1}}

\def\1{\bm{1}}

\DeclareMathAlphabet{\mathsfit}{\encodingdefault}{\sfdefault}{m}{sl}
\SetMathAlphabet{\mathsfit}{bold}{\encodingdefault}{\sfdefault}{bx}{n}

\newcommand{\method}
{\textsc{Method}}
\usepackage{hyperref}
\usepackage[x11names]{xcolor}
\usepackage{url}
\usepackage{natbib}
\usepackage[capitalize,noabbrev]{cleveref}
\usepackage{bm}
\usepackage{thm-restate}
\usepackage{enumitem}
\usepackage{algorithm}
\usepackage{newunicodechar}
\newunicodechar{★}{\ding{72}}
\usepackage{booktabs} 
\usepackage{tabularx} 
\usepackage{amssymb} 

\usepackage{pgfplots}
\usepackage{xcolor}
\definecolor{peach}{rgb}{1.0, 0.9, 0.71}
\usepackage[utf8]{inputenc}
\usepackage{pifont}
\usepackage{amsmath,amssymb,mathtools}
\usepackage{amsthm}

\newtheorem{lemma}{Lemma}

\newtheorem{definition}{Definition}
\usepackage{algorithm}                  
\usepackage[noend]{algpseudocode}       
\usepackage{microtype}
\usepackage{arydshln}
\usepackage{booktabs}
\usepackage{hyphenat}
\newcounter{assumpcounter}
\usepackage{hyperref}
\hypersetup{
    colorlinks=true,
    linkcolor=blue,
    filecolor=magenta,      
    urlcolor=cyan,
    citecolor=blue,
}
\usepackage{url} 
\usepackage{graphicx}
\usepackage[most]{tcolorbox}
\tcbset{
    boxsep=0mm,
    boxrule=0pt,
    colframe=white,
    arc=0mm,
    left=0.5mm,
    right=0.5mm
}
\usepackage{enumitem}
\usepackage{minitoc}
\usepackage{tikz}
\usetikzlibrary{arrows.meta,positioning,shapes.geometric,calc,fit,backgrounds}

\definecolor{agentblue}{RGB}{95,149,237}
\definecolor{envorange}{RGB}{255,183,77}
\definecolor{datablue}{RGB}{189,215,238}
\definecolor{warmgray}{RGB}{220,220,220}
\definecolor{outgreen}{RGB}{180,220,180}
\definecolor{diamondpeach}{RGB}
{247,231,206}
\definecolor{peach}{RGB}{247,231,206}
\usepackage{enumitem}
\usetikzlibrary{fadings}
\usetikzlibrary{mindmap,trees}
\usetikzlibrary{arrows,automata,shapes,positioning,shadows,trees}
\usetikzlibrary{decorations.pathmorphing}
\usetikzlibrary{shapes,shadows}
\usetikzlibrary{shapes.arrows}
\usetikzlibrary{calc,shapes, positioning}
\usetikzlibrary{decorations.text}
\usetikzlibrary{matrix,chains,positioning,decorations.pathreplacing,arrows}
\usetikzlibrary{bayesnet}
\usepackage[edges]{forest}
\usetikzlibrary{shadows.blur}
\usetikzlibrary{shapes.geometric}

\tikzstyle{edge}=[-latex',draw=black!90,shorten <=1pt,shorten >=1pt]
\tikzstyle{redge}=[latex'-,draw=black!90,shorten <=1pt,shorten >=1pt]
\tikzstyle{dedge}=[latex'-latex',draw=black!90,shorten <=1pt,shorten >=1pt]

\tikzstyle{block}=[draw, text width=5em,align=center,shape=rectangle, rounded corners, , align=center]
\tikzstyle{nobox}=[align=center]
\definecolor{emb}{RGB}{209,228,252}
\definecolor{hidden-blue}{RGB}{194,232,247}
\definecolor{hidden-orange}{RGB}{224,224,224}
\definecolor{hidden-yellow}{RGB}{242,244,193}
\definecolor{output-purple}{RGB}{219,203,231}
\definecolor{output-green}{RGB}{204,231,207}
\definecolor{output-blue}{RGB}{44,169,225}

\definecolor{output-black}{RGB}{0,0,0}
\definecolor{output-white}{RGB}{255,255,255}
\definecolor{myblue}{RGB}{137,195,235}

\definecolor{hiddendraw}{RGB}{137,195,235}

\tikzstyle{emb-purple}=[
    rectangle,
    draw=output-purple!50!purple,
    fill=output-purple,
    text opacity=1,
    minimum height=1.5em,
    minimum width=1.5em,
    inner sep=0pt,
    align=center,
    fill opacity=.5,
    ]

\tikzstyle{emb-blue}=[
    rectangle,
    draw=emb!50!blue,
    fill=emb,
    text opacity=1,
    minimum height=1.5em,
    minimum width=1.5em,
    inner sep=0pt,
    align=center,
    fill opacity=.5,
]

\newcommand{\headeright}{Preprint}

\newcommand{\shorttitle}{\@title}

\usepackage{fancyhdr}
\renewcommand{\headrulewidth}{0.4pt}
\fancyheadoffset{0pt}
\title{Competition is the key: A Game Theoretic Causal Discovery Approach}

\renewcommand{\shorttitle}{Competition is the key: A Game Theoretic Causal Discovery Approach}

\author{ \hspace{1mm}Amartya Roy\\
	The School of Interdisciplinary Research (SIRe),\\
	Indian Institute of Technology (IIT) Delhi,\\
	Hauz Khas - 110 016, New Delhi, India \\
    Robert Bosch GmbH, India\\
	\texttt{srz248670@iitd.ac.in} \\
	\And
	\hspace{1mm}Souvik~Chakraborty\\
    Department of Applied Mechanics,\\
    Yardi School of Artificial Intelligence, \\
    Indian Institute of Technology (IIT) Delhi,\\
    Hauz Khas - 110 016, New Delhi, India\\
    \texttt{souvik@am.iitd.ac.in} \\
}

\algrenewcommand\algorithmiccomment[1]{\hfill$\triangleright$~#1}

\begin{document}

\doparttoc 
\faketableofcontents 

\maketitle

\begin{abstract}
Causal discovery remains a central challenge in machine learning, yet existing methods face a fundamental gap: algorithms like GES and GraN-DAG achieve strong empirical performance but lack finite-sample guarantees, while theoretically principled approaches fail to scale. We close this gap by introducing a \textbf{game-theoretic reinforcement learning framework} for causal discovery, where a {DDQN} agent directly \emph{competes} against a strong baseline (GES or GraN-DAG), always \emph{warm-starting} from the opponent’s solution. This design yields \underline{three} provable guarantees: the learned graph is \emph{never worse} than the opponent, warm-starting \emph{strictly accelerates convergence}, and most importantly with high probability the algorithm selects the true best candidate graph. Formally, if the sample size $n$ is sufficiently large, specifically 
$
n \;\geq\; \tfrac{8L^2}{\Delta_n^2}\,\log\!\biggl(\tfrac{2|C|}{\delta}\biggr),
$
then with probability at least $1-\delta$ our method recovers the population-optimal graph. Here $L$ is a Lipschitz constant of the score function, $\Delta_n$ is the empirical gap between the best and second-best candidate scores, $|C|$ is the number of candidate graphs considered, and $ \forall\delta \in (0,1)$ is the failure probability . Thus, to the best of our knowledge, our result here makes a first-of-its-kind progress with explaining such finite-sample guarantees in causal discovery: on synthetic SEMs (30 nodes), the observed error probability decays with $n$, tightly matching theory. On real-world benchmarks--including Sachs, Asia, Alarm, Child, Hepar2, Dream, and Andes, our method consistently \emph{outperforms} GES and GraN-DAG while remaining \emph{theoretically safe}. Remarkably, it scales to \emph{large graphs} such as Hepar2 ($\sim$70 nodes), Dream ($\sim$100 nodes), and Andes ($\sim$220 nodes). Together, these results establish a {new class of RL-based causal discovery algorithms} that are simultaneously \emph{provably consistent}, \emph{sample-efficient}, and \emph{practically scalable}, marking a decisive step toward unifying \emph{empirical performance} with \emph{rigorous finite-sample theory}.
\end{abstract}

\section{Introduction}

Randomized controlled trials~\citep{hariton2018randomised} are widely regarded as the gold standard for causal inference, but in many domains they are infeasible, prohibitively expensive, or ethically questionable~\citep{chen2023role}. This limitation has driven sustained interest in causal discovery from observational data, yet every major family of algorithms comes with sharp drawbacks. Constraint-based methods such as PC~\citep{spirtes2001causation} and FCI~\citep{spirtes2001causation} rely on conditional independence tests, but suffer from instability: a single skeleton error can cascade into widespread orientation mistakes. Score-based methods like GES~\citep{chickering2002optimal} optimize likelihood criteria with complexity penalties, but the search is NP-hard, requiring greedy heuristics that can stall under finite samples or model misspecification. Functional causal models (e.g., LiNGAM~\citep{shimizu2014lingam}, ANM~\citep{hoyer2008nonlinear}) guarantee identifiability only under restrictive assumptions, and fail when real data violate them. Continuous optimization relaxations such as NOTEARS~\citep{zheng2018dags}, DAG-GNN~\citep{pmlr-v97-yu19a}, and GraN-DAG~\citep{lachapelle2019gradient} enforce acyclicity through smooth constraints, but are tied to specific surrogate losses, limiting their ability to incorporate arbitrary scores or robustness objectives.

Reinforcement learning (RL) has been proposed as a flexible paradigm for causal discovery. RL-BIC~\citep{zhu2020causal} showed that policy-based exploration can outperform GES on several benchmarks, and CORL~\citep{wang2021ordering} framed node ordering as a Markov decision process. More recently, KCRL~\citep{hasan2022kcrl} argued that prior knowledge can be injected via reward-penalty constraints to shrink the search space and accelerate convergence. Yet these methods remain essentially heuristic: RL-BIC exhibits unstable precision-recall trade-offs, CORL generalizes poorly beyond the Sachs dataset, and KCRL leaves open the fundamental question of whether reinforcement learning for causal discovery can be placed on firm theoretical ground.

\textbf{This work takes a step in addressing the above limitations.} We propose a Double Q-learning~ \citep{van2016deep} framework for causal discovery, \textsc{DDQN-CD}, that transforms RL-based search into a principled, theoretically controlled procedure. Our framework integrates robust BIC scores (Copula-BIC), warm-starts the search from strong classical opponents such as GES or GraN-DAG, and enforces feasibility through action masking and edge budgets. Crucially, the algorithm maintains a \emph{champion-challenger} setup: it never returns worse than its opponent, it provably reduces the expected time to reach a local optimum when warm-started, and it offers finite-sample guarantees that the probability of selecting a suboptimal graph decays exponentially with sample size. In short, we move RL-based causal discovery from heuristic exploration to a theoretically grounded optimization framework.

We validate our approach in two regimes. On synthetic data, we directly stress-test the theorem, demonstrating that the probability of mis-selection shrinks with $n$ while the population gap grows. On real benchmarks -Sachs~\citep{zhang2021gcastle}, Asia~\citep{lauritzen1988local}, Alarm~\citep{beinlich1989alarm}, Child~\citep{834c237a-5d0a-3651-a4ff-7a716db71a04}, Hepar2~\citep{onisko2003probabilistic}, DREAM~\citep{kalainathan2020causal}, and  Andes~\citep{conati1997line}
we demonstrate scalability. In particular, Hepar2, DREAM and Andes contain 70, 100, and 220 nodes respectively, where several competing RL-based or continuous methods fail outright, yet \textsc{DDQN-CD} consistently delivers competitive or superior structure recovery. These results establish \textsc{DDQN-CD} as a scalable, theoretically grounded alternative to existing causal discovery algorithms. Our contributions are summarised as follows( ref. Figure~\ref{fig:sum})

\paragraph{Organization.} 
The remainder of this paper is organized as follows. 
\Cref{sec:related} surveys prior work on causal discovery, with particular attention to reinforcement learning approaches. 
\Cref{sec:method} introduces our \textsc{DDQN-CD} framework, outlining the game-theoretic formulation, reward design, and learning algorithm. 
Theoretical guarantees are presented in detail in \Cref{sec:thm}. 
\Cref{sec:exp} provides both theoretical verification of \Cref{thm:champion} and empirical evaluations on benchmark datasets spanning small, mid-scale, and large networks, accompanied by insights and broader implications. 
Finally, \Cref{sec:conclusion} concludes the paper and highlights future research directions.

\begin{figure*}[htbp]
\label{fig:sum}
\centering
\scriptsize

\begin{minipage}[t]{0.32\textwidth}
\begin{tcolorbox}[
  title=Unified RL-based Framework,
  colback=blue!10, colframe=blue!80!black,
  fonttitle=\bfseries, boxrule=0.4pt, rounded corners,
  equal height group=contribs, left=1.5mm,right=1.5mm,top=1.5mm,bottom=1.5mm,
  valign=center
]
\begin{itemize}[leftmargin=*, itemsep=2pt, topsep=2pt, parsep=0pt, labelsep=0.4em]
  \item Double Q-learning for causal discovery
  \item Warm-start with GES / GraN-DAG
  \item Robust BIC (Copula-BIC)
\end{itemize}
\end{tcolorbox}
\end{minipage}
\hfill
\begin{minipage}[t]{0.32\textwidth}
\begin{tcolorbox}[
  title=Theoretical Guarantees,
  colback=orange!10, colframe=orange!70!black,
  fonttitle=\bfseries, boxrule=0.4pt, rounded corners,
  equal height group=contribs, left=1.5mm,right=1.5mm,top=1.5mm,bottom=1.5mm,
  valign=center
]
\begin{itemize}[leftmargin=*, itemsep=2pt, topsep=2pt, parsep=0pt, labelsep=0.4em]
  \item Never worse than the warm-start opponent
  \item Faster hitting time to a local optimum
  \item Suboptimal selection prob.\ decays exponentially in $n$
\end{itemize}
\end{tcolorbox}
\end{minipage}
\hfill
\begin{minipage}[t]{0.32\textwidth}
\begin{tcolorbox}[
  title=Scalability Across Real Datasets,
  colback=green!10, colframe=green!50!black,
  fonttitle=\bfseries, boxrule=0.4pt, rounded corners,
  equal height group=contribs, left=1.5mm,right=1.5mm,top=1.5mm,bottom=1.5mm,
  valign=center
]
\begin{itemize}[leftmargin=*, itemsep=2pt, topsep=2pt, parsep=0pt, labelsep=0.4em]
  \item \textbf{Small} (Asia, Sachs, Lucas): near-perfect (TPR$=1.0$, FDR$=0.0$ on Asia/Lucas)
  \item \textbf{Mid} (Alarm, Hepar2): balanced; stronger than RL-BIC2, competitive with GES
  \item \textbf{Large} (DREAM, Andes): SHD $\downarrow$ 30-40\% vs.\ Gran-DAG at 100-200+ nodes
\end{itemize}
\end{tcolorbox}
\end{minipage}

\caption{Summary of our contributions. \textsc{DDQN-CD} integrates a unified RL-based framework with theoretical guarantees and demonstrates scalability across diverse benchmarks.}
\label{fig:contribs}
\end{figure*}

\section{Related Work}
\label{sec:related}

Causal discovery has been extensively studied across multiple paradigms, including constraint-based approaches (e.g., PC~\citep{spirtes2001causation}), score-based search (e.g., GES~\citep{chickering2002optimal}), functional causal models such as LiNGAM~\citep{shimizu2006linear,shimizu2011directlingam}, and continuous optimization frameworks like NOTEARS~\citep{zheng2018dags}, GOLEM~\citep{ng2020role}, and GraN-DAG~\citep{lachapelle2019gradient}. While these methods provide strong theoretical guarantees or computational elegance, they often struggle with scalability, robustness to noise, or the ability to balance precision and recall in large networks.

Reinforcement learning (RL) has emerged as a flexible alternative, framing causal discovery as a sequential decision-making problem. RL-BIC2~\citep{zhu2020causal} introduced actor-critic based search guided by BIC rewards, but it suffers from instability and limited recall. CORL~\citep{wang2021ordering} cast node ordering as an MDP but is restricted in applicability beyond small datasets such as Sachs. More recently, KCRL~\citep{hasan2022kcrl} incorporated domain knowledge via reward-penalty shaping, improving convergence but leaving open questions of scalability. These works highlight both the promise and the limitations of RL-based discovery. In contrast, our method leverages Double DQN with opponent warm starts (GES or GraN-DAG) and BIC-based rewards, transforming RL from a heuristic into a scalable, theoretically grounded framework that consistently outperforms across small, mid, and large networks.

\section{Method: Game Theoretic Causal Discovery}
\label{sec:method}

\begin{figure}[t]
\centering
\small
\resizebox{\columnwidth}{!}{%
\begin{tikzpicture}[
  font=\scriptsize,
  node distance=4mm and 8mm,
  box/.style={draw, rounded corners=2pt, align=center,
              minimum height=0.95cm, semithick, inner sep=3pt, fill=white},
  input/.style={box, text width=3.1cm, fill=datablue!18},
  agent/.style={draw, rounded corners=3pt, align=center,
                minimum height=1.15cm, semithick, inner sep=3pt,
                fill=output-purple!22, text width=4.1cm},
  env/.style={box, fill=envorange!18, text width=4.2cm},
  decision/.style={diamond, draw, semithick, aspect=1.8, align=center,
                   inner sep=1pt, minimum height=1.05cm, minimum width=1.65cm, fill=peach!35},
  flow/.style={-{Latex[length=2mm]}, semithick},
  feedback/.style={-{Latex[length=2mm]}, semithick, dashed},
  lane/.style={rounded corners=4pt, fill=black!3, draw=black!6, line width=0.2pt},
  merge/.style={circle, draw, fill=white, inner sep=0pt, minimum size=2.0mm}
]
\node[input] (data) {Observational Data\\$\displaystyle X\!\in\!\mathbb{R}^{n\times p}$};
\node[input, below=3mm of data] (warm)
  {Opponent init.\ (GES / GraN\text{-}DAG)\\\footnotesize binarized $A_0$};
\path (data.east) -- (warm.east) coordinate[pos=0.56] (mid);
\node[merge, right=6mm of mid] (m) {};

\node[agent, right=14mm of data, yshift=-1mm] (agent)
  {\textbf{DDQN Agent}\\[-1pt]\footnotesize Actions: add / remove / reverse};

\node[env, below=6mm of agent] (env)
  {\textbf{RL Environment}\\[-1pt]
   \footnotesize Score: BIC / Copula\mbox{-}BIC\\[-1pt]
   \footnotesize Reward: $\displaystyle r=\tfrac{\Delta S}{p}-\lambda\lVert A'\rVert_0-c$};

\node[decision, right=14mm of agent] (stop) {Stop?};
\node[box, fill=green!25, below=5mm of stop, text width=2.35cm, align=center] (out)
  {\textbf{Discovered DAG} $\hat G$\\\footnotesize (optional CAM)};

\draw[flow] (data.east) .. controls +(6mm,0) and +(-5mm,7mm) .. (m);
\draw[flow] (warm.east) .. controls +(6mm,0) and +(-5mm,-7mm) .. (m);
\draw[flow] (m) -- (agent.west);

\draw[flow] (agent.south) -- node[right, xshift=1mm] {\footnotesize proposed $A'$} (env.north);
\draw[flow] (agent.east) -- ++(4mm,0) -- (stop.west);

\draw[flow] (stop.south) -- node[right] {\footnotesize Yes} (out.north);

\draw[-{Latex[length=2mm]}, semithick]
  (env.east) to[out=8, in=-65]
  node[pos=0.55, fill=white, inner sep=1pt] {\footnotesize reward}
  (agent.east);

\draw[feedback]
  (stop.north) to[out=115, in=65, looseness=1.0]
  node[pos=0.50, fill=white, inner sep=0.8pt] {\footnotesize No}
  (agent.north);

\begin{scope}[on background layer]
  \node[lane, fit=(data) (warm), inner sep=5pt] (laneL) {};
  \node[lane, fit=(agent) (env), inner sep=7pt] (laneC) {};
  \node[lane, fit=(stop) (out),  inner sep=7pt] (laneR) {};
\end{scope}
\node[anchor=west, font=\footnotesize\itshape, text=black!65]
  at ([yshift=1.2mm]laneL.north west) {Input \& Warm Start};
\node[anchor=west, font=\footnotesize\itshape, text=black!65]
  at ([yshift=1.2mm]laneC.north west) {Agent \& Environment};
\node[anchor=west, font=\footnotesize\itshape, text=black!65]
  at ([yshift=1.2mm]laneR.north west) {Decision \& DAG};
\end{tikzpicture}%
}
\caption{\textsc{DDQN--CD} framework. Observational data and a warm start are merged and fed to a DDQN agent that proposes edge edits; an environment evaluates candidates via BIC/Copula\mbox{-}BIC and returns a reward until the stopping condition triggers, yielding the discovered DAG.}
\label{fig:framework}
\end{figure}
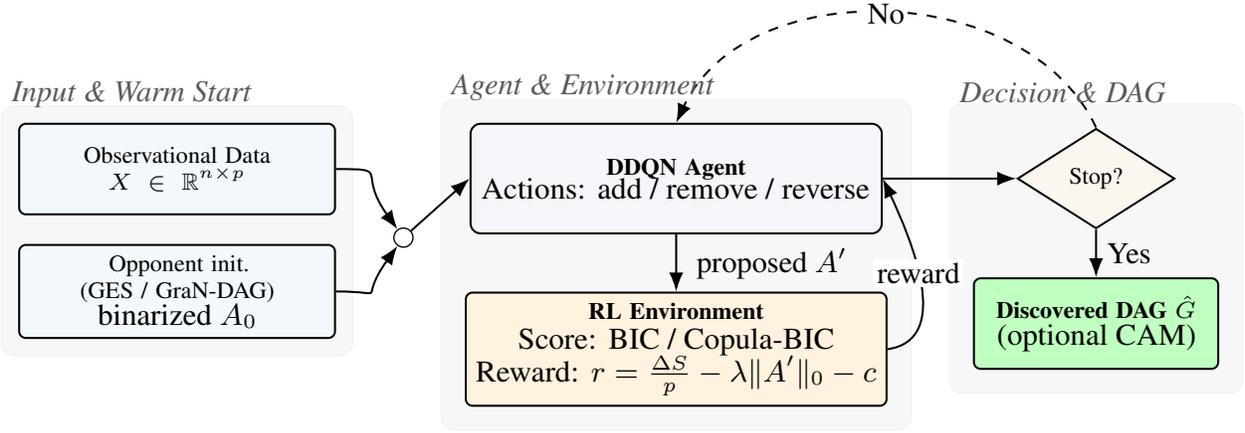

We cast causal discovery as a \emph{sequential game} between a reinforcement learning agent and an opponent prior (GES or GraN-DAG), which provides a warm-start graph $A_0$. The agent refines $A_0$ through local edge edits (\textsc{Add}, \textsc{Remove}, \textsc{Reverse}), restricted to acyclicity and edge-budget constraints. Each move receives a payoff
\[
r(A \!\to\! A') \;=\; \tfrac{S(A') - S(A)}{p} - \lambda \|A'\|_0 - c ,
\]
capturing normalized BIC improvement, sparsity, and step cost. Training proceeds via Double DQN with replay buffer and Polyak updates, where the agent selects actions $\varepsilon$-greedily until a stopping criterion is met (Algo.~\ref{alg:dqncd}). This \emph{champion--challenger} setup guarantees the discovered DAG $\widehat{G}$ is never worse than its opponent, turning strong priors into stepping stones for scalable, accurate causal discovery. {\color{violet}  For more details, please refer \cref{app:method}}

\begin{algorithm}[h]
\small
\caption{DDQN\textendash CD: Double Deep Q\textendash Learning for Causal Discovery with BIC Reward and Opponent Warm Start}
\label{alg:dqncd}
\begin{algorithmic}[1]
\Require Data $X\!\in\!\mathbb{R}^{n\times p}$; opponent flag $o\!\in\!\{\mathrm{GraN\text{-}DAG},\mathrm{GES}\}$; hyperparameters $(\gamma,\tau,\lambda,c,B,T,E,P)$
\Ensure Discovered DAG $\widehat G$
\State \textbf{Scorer} $S$: use \emph{DiscreteBIC} if $X$ is binary, else \emph{CopulaBIC} (Gaussian BIC after rank–Gaussian transform)
\State \textbf{Warm start} $A_0 \gets \mathrm{binarize}(\mathrm{Opponent}(X,o))$ \Comment{GraN\text{-}DAG or GES output}
\State \textbf{Actions} on ordered pairs $(i,j),\, i\neq j$: \textsc{Add}$(i\!\to\!j)$, \textsc{Remove}$(i\!\to\!j)$, \textsc{Reverse}$(i\!\to\!j)$
\State Mask invalid actions to keep acyclicity and edge budget $\|A\|_0 \le B$
\State \textbf{Reward} for $A\!\to\!A'$:
       \[
         r(A\!\to\!A') \;=\; \frac{S(A') - S(A)}{p} \;-\; \lambda\,\|A'\|_0 \;-\; c
       \]
\State Initialize online $Q_\theta$ and target $\bar Q_{\bar\theta}\!\leftarrow\!Q_\theta$; replay buffer $\mathcal{M}$; best graph $\widehat G\!\leftarrow\!A_0$
\For{$e=1$ \textbf{to} $E$} \Comment{episodes}
  \State $A\leftarrow A_0$ \Comment{reset}
  \For{$t=1$ \textbf{to} $T$} \Comment{steps}
    \State $m \leftarrow$ valid action mask from $A$
    \State \textbf{(action)} with prob.\ $\varepsilon$: sample $a$ uniformly from $\{k : m_k=1\}$; else $a \leftarrow \arg\max_{k:\, m_k=1} Q_\theta(A,k)$
    \State \textbf{(transition)} If $a$ valid \& keeps DAG/budget produce $A'$; otherwise set $A'\!\leftarrow\!A$, reward $r\!\leftarrow\!-\delta$
    \State \textbf{(reward)} If valid, set $r \leftarrow \frac{S(A')-S(A)}{p}-\lambda\|A'\|_0-c$
    \State Store $(A,a,r,A')$ in $\mathcal{M}$; set $A\leftarrow A'$
    \If{$|\mathcal{M}|$ sufficient} \Comment{Double DQN update}
        \State Sample mini\hyp batch $\{(A^i,a^i,r^i,{A'}^i)\}_{i=1}^b$ from $\mathcal{M}$
        \State $y^i \leftarrow r^i + \gamma\, \bar Q_{\bar\theta}\!\big({A'}^i,\arg\max_k Q_\theta({A'}^i,k)\big)$
        \State Update $\theta$ by one SGD step on $\frac{1}{b}\sum_i \big(Q_\theta(A^i,a^i)-y^i\big)^2$
        \State Polyak target: $\bar\theta \leftarrow (1-\tau)\bar\theta + \tau\,\theta$
    \EndIf
  \EndFor
  \If{$e \bmod P = 0$ \textbf{and} $S(A) > S(\widehat G)$} \; $\widehat G \leftarrow A$ \EndIf
\EndFor
\State \Return $\widehat G$ \Comment{(Optionally apply CAM pruning post hoc)}
\end{algorithmic}
\end{algorithm}

\section{Theoretical Guarantees}
\label{sec:thm}
\subsubsection*{Preliminaries}

Let $p$ be the number of variables. The finite set of actions $\mathcal{A}\subset\{0,1\}^{p\times p}$ consists of binary adjacency matrices respecting acyclicity and a configured edge budget $B$. An action is one of $\{\text{add}(i\!\to\!j),\text{remove}(i\!\to\!j),\text{reverse}(i\!\to\!j)\}$ for $i\neq j$. Each episode resets to a warm start graph $\tilde G$ and consists of at most $L$ edits.

Given the full dataset $X_{1:n}$, the empirical score is the Bayesian Information Criterion (BIC):
\[
S_n(A)\;=\;\sum_{t=1}^n \ell(A;X_t)\;-\;\frac{1}{2}\,k(A)\,\log n,
\]
where $\ell$ is the log-likelihood per sample and $k(A)$ is the parameter count. The algorithm maintains a \emph{champion} snapshot $\widehat G$ with the largest score seen and ultimately returns the better of $\widehat G$ and the opponent $\tilde G$. All the necessary assumptions are mentioned in Table~\ref{tab:assumptions}

\begin{table}[htbp]
\centering
\small
\caption{Summary of Theoretical Assumptions}
\label{tab:assumptions}
\setcounter{assumpcounter}{0} 
\begin{tabularx}{\textwidth}{@{} >{\bfseries}l X @{}}
\toprule
\multicolumn{1}{l}{\textbf{Assumption}} & \textbf{Formal Description} \\
\midrule

\refstepcounter{assumpcounter}\label{as:finite}
A\arabic{assumpcounter}: Finite Feasibility & 
Acyclicity and the edge budget $B$ are enforced by masking, so the feasible set $\mathcal{A}$ is finite. Each episode lasts at most $L<\infty$ steps. \\ \addlinespace

\refstepcounter{assumpcounter}\label{as:eps}
A\arabic{assumpcounter}: Persistent Exploration & 
The policy explores with probability at least $\varepsilon_\star>0$. When exploring, the action is chosen uniformly from all valid actions. \\ \addlinespace

\refstepcounter{assumpcounter}\label{as:warm}
A\arabic{assumpcounter}: Warm Start & 
Every episode initializes at the opponent DAG $A_0=\tilde G$. \\ \addlinespace

\refstepcounter{assumpcounter}\label{as:savebest}
A\arabic{assumpcounter}: Champion-Challenger & 
The algorithm returns $G_{\text{out}} = \arg\max\big\{\,S_n(\widehat G),\ S_n(\tilde G)\,\big\}$, where $\widehat G$ is the best graph found by the agent. \\ \addlinespace

\refstepcounter{assumpcounter}\label{as:std-gauss}
A\arabic{assumpcounter}: Gaussian Data & 
After preprocessing, observations $Z_1,\dots,Z_n \in \mathbb R^p$ are i.i.d.\ from $\mathcal N(0,I_p)$. \\ \addlinespace

\refstepcounter{assumpcounter}\label{as:lipschitz-L}
A\arabic{assumpcounter}: Lipschitz Score & 
The score can be written as $S_n(A)\;=\;\sum_{t=1}^n s_A(Z_t)\;-\;\tfrac{1}{2}k(A)\log n$, where $s_A$ is $L$-Lipschitz w.r.t.\ $\|\cdot\|_2$ i.e
$|s_A(x)-s_A(y)|\le L\|x-y\|_2$ $\forall$ $x,y$. \\

\bottomrule
\end{tabularx}
\end{table}

\subsubsection*{Guarantee I: Safety}

\begin{tcolorbox}
\begin{restatable}[Never worse than opponent]{theorem}{safety} \label{thm:safety}
Under (A\ref{as:savebest}), the returned graph $G_{\text{out}}$ satisfies $S_n(G_{\text{out}})\ \ge\ S_n(\tilde G)$.
\end{restatable}
\end{tcolorbox}
\begin{proof}(Sketch; a complete proof is in Appendix~ \ref{app:proofs1})\small.
By (A\ref{as:savebest}), the algorithm outputs the maximizer between the incumbent $\widehat G$ and the opponent $\tilde G$. The incumbent, by definition, has a score at least as high as any graph the agent visited. The inequality holds identically.
\end{proof}

\subsubsection*{Guarantee II: Efficient Exploration from Warm-Start}

\begin{definition}[1-optimal DAG]
A DAG $G^\star\in\mathcal{A}$ is \emph{1-optimal} if no valid single-edge edit $e$ improves the score, i.e., $S_n(G^\star\oplus e)\le S_n(G^\star)$.
\end{definition}
\begin{tcolorbox}
\begin{restatable}[Hitting time bound]{theorem}{hitting}\label{thm:hitting}
Under (A\ref{as:finite})-(A\ref{as:warm}), if the episode horizon $L$ is long enough to reach a 1-optimal $G^\star$ from the warm-start $\tilde G$, the expected number of episodes $\mathbb{E}[T]$ to visit $G^\star$ is bounded: $\mathbb{E}[T]\ \le\ (\varepsilon_\star/A_{\max})^{-d(\tilde G,G^\star)}$, where $d(\cdot,\cdot)$ is the shortest improving path length and $A_{\max}$ is the maximum number of valid actions. A better warm start (smaller $d$) geometrically improves the bound.
\end{restatable}
\end{tcolorbox}

\begin{proof}(Sketch; a complete proof is in Appendix~ \ref{app:proofs2})\small. \,
By Lemma \ref{lem:path}, a strictly improving path of length $d=d(\tilde G,G^\star)$ exists. In any episode, the probability of following this specific path via exploration is at least $\pi_{\min} = (\varepsilon_\star/A_{\max})^d$. Let $I_e$ be the indicator that episode $e$ hits $G^\star$. Then $\mathbb{P}(I_e=1)\ge \pi_{\min}$. The number of episodes $T$ until the first hit is therefore stochastically dominated by a Geometric($\pi_{\min}$) random variable, whose expectation is $1/\pi_{\min}$.
\end{proof}

\subsubsection*{Guarantee III: Finite-Sample Champion Selection}

\begin{tcolorbox}
\begin{restatable}[High-probability champion selection]{theorem}{champion}
\label{thm:champion}
Let $\mathcal C$ be the set of candidate graphs (agent snapshots and $\tilde G$). Let $A_n^\diamond$ be the unique graph that maximizes the population-level score. Under (A\ref{as:std-gauss})-(A\ref{as:lipschitz-L}), for any $\delta\in(0,1)$, if the sample size $n$ is sufficiently large, i.e., $n\ \ge\ \frac{8\,L^2}{\Delta_n^2}\,\log(\frac{2|\mathcal C|}{\delta})$, then the graph returned by the algorithm is the true best candidate with probability at least $1-\delta$. Here, $\Delta_n$ is the gap between the best and second-best candidate scores.
\end{restatable}
\end{tcolorbox}

\begin{proof}(Sketch; a complete proof is in Appendix~ \ref{app:proofs3})\small. \, Let $G_{\mathrm{out}}$ be the returned graph. The event $G_{\mathrm{out}}\neq A_n^\diamond$ implies that for some other graph $A$, $S_n(A) \ge S_n(A_n^\diamond)$. We can bound the probability of this error for a fixed $A$. The difference in scores, $D_n(A) = S_n(A_n^\diamond)-S_n(A)$, is a sum of i.i.d. random variables. By demonstrating that the terms are sub-Gaussian (via the Lipschitz property), we can apply a Chernoff bound to show that $\mathbb P(D_n(A)\le 0)\le \exp(-n\Delta_n^2/(8L^2))$. A union bound over all other candidates in $\mathcal C$ gives the final result.
\end{proof}

\subsubsection*{Takeaway}
Under assumptions (A1)–(A4), our procedure is (i) \textit {safe}, never performing worse than the opponent (Thm. \ref{thm:safety}) (ii) \textit {efficient}, reaching a 1-optimal DAG in geometrically fewer episodes from a warm start (Thm. \ref{thm:hitting}) and (iii) \textit {consistent}, selecting the best overall candidate with high probability given enough data (Thm. \ref{thm:champion}).

\section{Experiments}
\label{sec:exp}

\subsection{Verifying \textbf{Theorem} ~\ref{thm:champion} on Synthetic Data}

\begin{figure}[h]
    \centering
\includegraphics[width=0.45\linewidth]{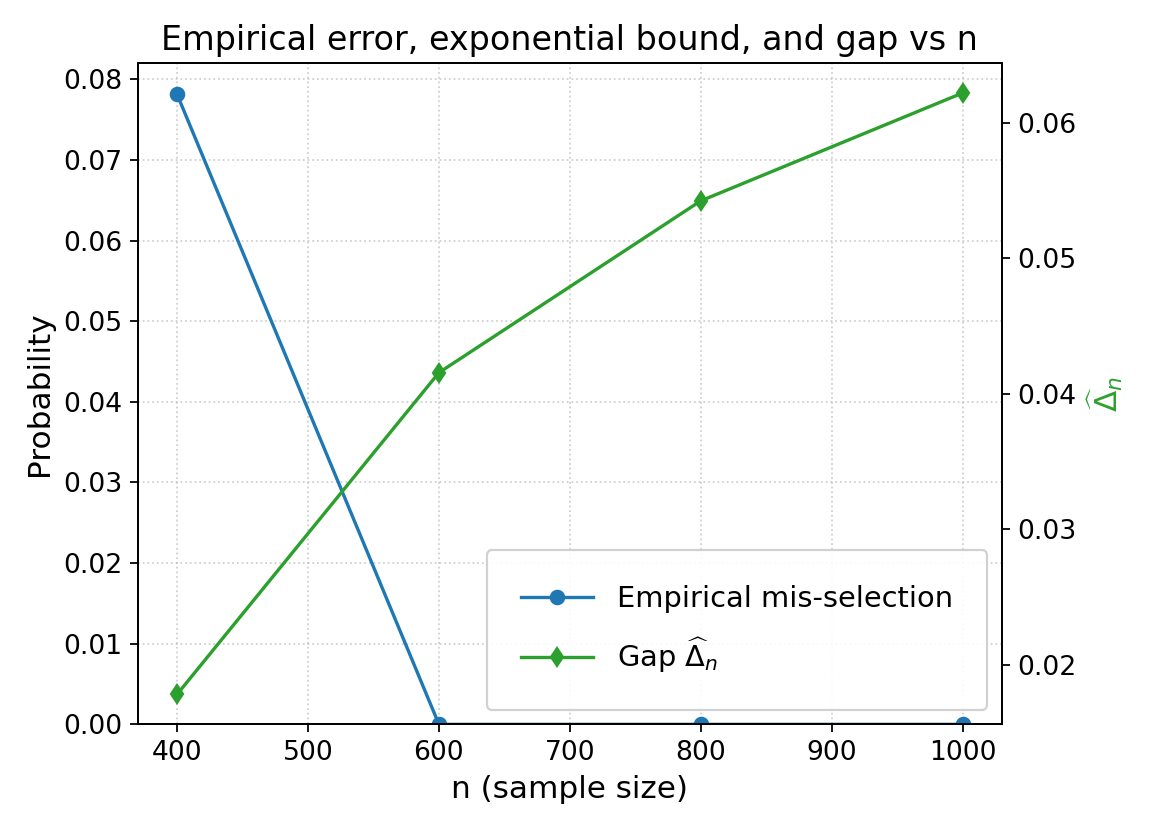}
\caption{Synthetic verification of Theorem~\ref{thm:champion}. 
    The mis-selection probability (blue, left axis) decays with $n$, while the gap $\Delta_n$ (green, right axis) grows with $n$, matching theoretical predictions.}
    \label{fig:synthetic-verify}
\end{figure}

The synthetic study serves as a direct verification of our finite-sample selection theorem (Theorem~\ref{thm:champion}).  
The theorem states that (i) the probability of mis-selecting a suboptimal candidate from the fixed set $\mathcal{C}$ decreases exponentially with sample size $n$ and (ii) the per-sample population gap $\Delta_n$ between the best and second-best candidates increases with $n$ as the penalty term vanishes.  
\textit {Setup}:   
We constructed synthetic data from a linear-Gaussian SEM with $p=30$ nodes. A random DAG was sampled with expected in-degree $3$, and edge weights were drawn uniformly in $[0.5,1.0]$ with random sign. Observations were generated as $X=(I-W)^{-1}e$ with $e\sim\mathcal{N}(0,I)$.  
The data was split into a training (for candidate construction) and a validation pool (for estimating population quantities).  
The candidate set $\mathcal{C}$ was built by combining (a) the opponent structure from GES, (b) our DDQN agent starting from the opponent, and (c) CAM-pruned refinements. This set was fixed across sample sizes.  
\textit {Evaluation}: 
For each $n \in \{400,600,800,1000\}$, we repeated $40$ independent trials. In each trial we drew $n$ new samples, computed empirical BIC scores $S_n(A)$ for $A\in\mathcal{C}$, and returned $G_{\text{out}}=\arg\max_{A\in\mathcal{C}} S_n(A)$.  
We compared $G_{\text{out}}$ with the population best $A_n^\diamond=\arg\max_{A\in\mathcal{C}} \Lambda_n(A)$, where $\Lambda_n(A)=\mu(A)-\tfrac{k(A)}{2n}\log n$ was estimated from the validation pool.  
This allowed us to estimate both the mis-selection probability $\mathbb P(G_{\text{out}}\neq A_n^\diamond)$ and the gap $\Delta_n$.  
\textit {Results}:  
Figure~\ref{fig:synthetic-verify} summarizes the findings. The blue curve shows that the empirical error probability falls rapidly with $n$, reaching near zero by $n=600$. The green curve shows that the population gap $\Delta_n$ increases with $n$, consistent with the theorem.  
Together, these results confirm that with more samples, the chance of selecting a suboptimal candidate decreases exponentially, while the effective separation between the best and second-best graphs widens (Theorem ~\ref{thm:champion}).   

\subsection{Baselines}
To ensure a comprehensive and fair evaluation, we compare our framework against a diverse set of \textit{state-of-the-art} causal discovery methods spanning constraint-based, score-based, functional, gradient-based, and reinforcement learning paradigms. This diversity ensures that our benchmarks reflect both classical and modern advances in the field.
\textit {Constraint-based methods} $\Rightarrow$
The PC algorithm~\citep{spirtes2001causation} identifies structures using conditional independence tests, while FCI extends PC to handle latent confounding. These approaches are computationally efficient on small graphs but often unstable under sampling noise, where small skeleton errors propagate into widespread orientation mistakes.
\textit {Score-based methods} $\Rightarrow$
GES~\citep{chickering2002optimal} remains the most widely used representative, employing greedy equivalence search with BIC scoring. It is robust on moderately sized networks but relies on NP-hard optimization, limiting its scalability.  
\textit {Functional causal models} $\Rightarrow$
We include LiNGAM~\citep{shimizu2006linear} and DirectLiNGAM~\citep{shimizu2011directlingam}, which exploit non-Gaussianity to guarantee identifiability. ICALiNGAM~\citep{shimizu2006linear} extends this principle via ICA. While theoretically elegant, these models are brittle under model misspecification.
\textit {Gradient-based optimization} $\Rightarrow$
Continuous optimization methods relax acyclicity into smooth constraints. NOTEARS~\citep{zheng2018dags} introduced the differentiable acyclicity constraint, later extended in DAG-GNN and GOLEM~\citep{ng2020role}. GraN-DAG~\citep{lachapelle2019gradient} further employs gradient-based generative modeling. These methods are elegant but prone to overfitting or collapse in large graphs. \textit {Reinforcement learning baselines} $\Rightarrow$
RL-BIC2~\citep{zhu2020causal} learns causal structures by optimizing BIC-guided rewards through reinforcement learning, but suffers from instability and poor scalability. CORL~\citep{wang2021ordering} formulates node ordering as an MDP but is tailored only for the Sachs dataset, failing to generalize. More recently, KCRL~\citep{hasan2022kcrl} incorporates prior knowledge through reward-penalty shaping, narrowing the search space. These RL-based methods validate the promise of learning-based search but lack generality across scales. \emph{By evaluating against all of them, we validate that our framework is not narrowly tuned but rather competitive across methodological families.}
\subsection{Real Datasets: Proving Scalability}
\begin{table*}[htbp]
\small
\centering
\caption{Performance comparison of different models across various metrics on \method. We highlight the best (\textbf{bold}) and second-best (\underline{underline}) values.
Columns labeled [(\(\uparrow\))] indicate higher-is-better; columns labeled [(\(\downarrow\))] indicate lower-is-better. All numeric values are rounded to two decimal places.}
\label{tbl:model_results_no_f1}

\begin{tabular}{lcccccccc}
\toprule
\multicolumn{9}{c}{\textbf{Small}} \\
\hdashline\\[-1.5ex]
& \multicolumn{4}{c}{\textbf{Asia}}
& \multicolumn{4}{c}{\textbf{Sach}} \\
\cmidrule(lr){2-5}\cmidrule(lr){6-9}
\textbf{Model Name} &
\textbf{TPR}\(\uparrow\) & \textbf{FDR}\(\downarrow\) & \textbf{SHD}\(\downarrow\) & \textbf{Score}\(\uparrow\) &
\textbf{TPR}\(\uparrow\) & \textbf{FDR}\(\downarrow\) & \textbf{SHD}\(\downarrow\) & \textbf{Score}\(\uparrow\) \\
\midrule
KCRL & 0.55 & 0.25 & \underline{3} & 0.52 & 0.35 & \underline{0.45} & 15 & 0.32 \\
NOTEARS & 0.13 & 0.83 & 12 & 0.13 & 0.30 & 0.59 & \textbf{12} & 0.26 \\
GOLEM & 0.25 & 0.75 & 11 & 0.19 & 0.18 & 0.83 & 24 & 0.13 \\
RL-BIC & 0.53 & 0.55 & 7 & 0.37 & 0.24 & 0.67 & \underline{14} & 0.21 \\
ICALiNGAM & 0.25 & 0.60 & 7 & 0.26 & 0.22 & 0.50 & \underline{14} & 0.26 \\
DirectLiNGAM & 0.50 & \textbf{0.00} & 4 & \underline{0.57} & 0.12 & 0.50 & 15 & 0.23 \\
PC & \underline{0.75} & 0.33 & 4 & 0.54 & 0.33 & 0.77 & 30 & 0.20 \\
CORL & NA& NA& NA& NA& 0.77 & 0.77 & 26 & 0.18 \\
Gran-DAG & 0.13 & \underline{0.13} & 7 & 0.42 & 0.15 & \textbf{0.30} & 15 & 0.30 \\
GES & \textbf{1.00} & \textbf{0.00} & \textbf{0} & \textbf{1.00} & \underline{0.78} & 0.64 & 28 & \underline{0.39} \\

\textbf{Ours(Using Gran-DAG)} & 0.63 & 0.38 & 5 & 0.47 & 0.45 & 0.60 & 18 & 0.30 \\
\textbf{Ours (using GES)}& \textbf{1.00}& \textbf{0.00}& \textbf{0}& \textbf{1.00}& \textbf{0.80} & 0.62 & 28 & \textbf{0.40}  \\
\bottomrule
\end{tabular}

\vspace{1em} 

\begin{tabular}{lcccccccc}
\toprule
\multicolumn{9}{c}{\textbf{Small (continued)}} \\
\hdashline\\[-1.5ex]
& \multicolumn{4}{c}{\textbf{Lucas}}
& \multicolumn{4}{c}{\textbf{Child}} \\
\cmidrule(lr){2-5}\cmidrule(lr){6-9}
\textbf{Model Name} &
\textbf{TPR}\(\uparrow\) & \textbf{FDR}\(\downarrow\) & \textbf{SHD}\(\downarrow\) & \textbf{Score}\(\uparrow\) &
\textbf{TPR}\(\uparrow\) & \textbf{FDR}\(\downarrow\) & \textbf{SHD}\(\downarrow\) & \textbf{Score}\(\uparrow\) \\
\midrule
KCRL & 0.36 & 0.43 & 8 & 0.35 & 0.15 & 0.80 & 28 & 0.13 \\
NOTEARS & 0.33 & 0.43 & 11 & 0.33 & 0.12 & 0.62 & \underline{22} & 0.18 \\
GOLEM & 0.45 & 0.50 & 9 & 0.35 & 0.10 & 0.78 & 24 & 0.12 \\
RL-BIC & 0.36 & 0.67 & 11 & 0.26 & \underline{0.44} & \textbf{0.39} & \textbf{21} & \underline{0.36} \\
ICALiNGAM & 0.18 & 0.67 & 10 & 0.20 & 0.24 & \underline{0.54} & \textbf{21} & 0.24 \\
DirectLiNGAM & 0.36 & 0.50 & 8 & 0.32 & 0.12 & 0.82 & 28 & 0.11 \\
PC & \underline{0.92} & \underline{0.08} & \underline{2} & \underline{0.72} & 0.24 & 0.86 & 43 & 0.13 \\
Gran-DAG & 0.09 & 0.50 & 10 & 0.23 & 0.50& 0.67& 25& 0.29 \\
GES & \textbf{1.00} & \textbf{0.00} & \textbf{0} & \textbf{1.00} & 0.38 & 0.89 & 34 & 0.17 \\
\textbf{Ours(Using Gran-DAG)}  & 0.33 & 0.83 & 25 & 0.18 & \textbf{0.72} & 0.55 & \textbf{21} & \textbf{0.41} \\
\textbf{Ours (using GES)} & \textbf{1.00} & \textbf{0.00} & \textbf{0} & \textbf{1.00} & 0.32 & 0.81 & 33 & 0.18 \\
\bottomrule
\end{tabular}

\vspace{1em} 

\begin{tabular}{lcccccccc}
\toprule
\multicolumn{9}{c}{\textbf{Mid}} \\
\hdashline\\[-1.5ex]
& \multicolumn{4}{c}{\textbf{Alarm}}
& \multicolumn{4}{c}{\textbf{Hepar2}} \\
\cmidrule(lr){2-5}\cmidrule(lr){6-9}
\textbf{Model Name} &
\textbf{TPR}\(\uparrow\) & \textbf{FDR}\(\downarrow\) & \textbf{SHD}\(\downarrow\) & \textbf{Score}\(\uparrow\) &
\textbf{TPR}\(\uparrow\) & \textbf{FDR}\(\downarrow\) & \textbf{SHD}\(\downarrow\) & \textbf{Score}\(\uparrow\) \\
\midrule
KCRL & 0.33 & 0.63 & 49 & 0.24 & NA& NA&NA &NA \\
NOTEARS & 0.17 & 0.43 & 41 & 0.26 & 0.02 & 0.99 & 157 & 0.01 \\
GOLEM & 0.15 & \underline{0.40} & 43 & 0.26 & NA& NA& NA& NA\\
RL-BIC & 0.30 & 0.74 & 56 & 0.20 & NA& NA& NA& NA\\
ICALiNGAM & 0.57 & \textbf{0.32} & \textbf{29} & \underline{0.42} & 0.19 & 0.49 & 112 & 0.24 \\
DirectLiNGAM & 0.39 & 0.50 & \underline{40} & 0.30 & 0.10 & \textbf{0.07} & 110 & 0.35 \\
PC & 0.67 & 0.60 & 55 & 0.36 & 0.35 & 0.75 & 172 & 0.20 \\
Gran-DAG & 0.24 & 0.73 & 60 & 0.18 & 0.28 & 0.39 & 96 & 0.30 \\
GES & \underline{0.74} & 0.61 & 56 & 0.38 & 0.50 & \underline{0.23} & \textbf{70} & \underline{0.42} \\
\textbf{Ours(Using Gran-DAG)}  & 0.40 & 0.65 & 49 & 0.26 & \textbf{0.54} & 0.51 & \underline{84} & 0.35 \\
\textbf{Ours (using GES)} & \textbf{0.82} & 0.55 & 58 & \textbf{0.43} & \underline{0.52} & \underline{0.23} & \textbf{70} & \textbf{0.43} \\
\bottomrule
\end{tabular}

\vspace{1em} 

\begin{tabular}{lcccccccc}
\toprule
\multicolumn{9}{c}{\textbf{Large}} \\
\hdashline\\[-1.5ex]
& \multicolumn{4}{c}{\textbf{Dream}}
& \multicolumn{4}{c}{\textbf{Andes}} \\
\cmidrule(lr){2-5}\cmidrule(lr){6-9}
\textbf{Model Name} &
\textbf{TPR}\(\uparrow\) & \textbf{FDR}\(\downarrow\) & \textbf{SHD}\(\downarrow\) & \textbf{Score}\(\uparrow\) &
\textbf{TPR}\(\uparrow\) & \textbf{FDR}\(\downarrow\) & \textbf{SHD}\(\downarrow\) & \textbf{Score}\(\uparrow\) \\
\midrule
NOTEARS & 0.07 & \underline{0.97} & 293 & 0.03 & 0.07 & \underline{0.97} & 316 & 0.03 \\
Gran-DAG & \underline{0.09} & \underline{0.97} & \underline{251} & \underline{0.04} & \underline{0.09} & \underline{0.97} & \underline{314} & \underline{0.04} \\
\textbf{Ours(Using Gran-DAG)}  & \textbf{0.15} & \textbf{0.82} & \textbf{184} & \textbf{0.11} & \textbf{0.12} & \textbf{0.89} & \textbf{284} & \textbf{0.08} \\
\bottomrule
\end{tabular}
\end{table*}
To demonstrate the scalability and robustness of our framework, we benchmarked on a suite of widely used real-world causal discovery datasets (Appendix~\ref{datasets}), ranging from small-scale networks such as Asia (8 nodes) and Sachs (11 nodes) to mid-sized graphs like Alarm (37 nodes) and Hepar2 (70+ nodes), and large-scale networks including Dream1 (100 nodes) and Andes (223 nodes). Table~\ref{tbl:model_results_no_f1} summarizes the performance across four metrics: \emph{True Positive Rate (TPR)}, \emph{False Discovery Rate (FDR)}, \emph{Structural Hamming Distance (SHD)}, and a composite \emph{Score}, defined to understand the overall performance. 

\vspace{0.5em}
\noindent\textbf{Composite Score.}  
Evaluating causal discovery methods typically involves reporting multiple metrics, most commonly true positive rate (TPR), false discovery rate (FDR), and structural Hamming distance (SHD). Each of these captures a different aspect of performance: TPR measures the ability to recover true edges, FDR quantifies spurious discoveries, and SHD reflects overall structural accuracy. However, these metrics can sometimes paint an incomplete or even conflicting picture. For instance, a method that is overly conservative may achieve a low SHD by predicting very few edges, but this comes at the cost of a poor TPR. Conversely, a method that aggressively predicts edges may achieve higher TPR but suffer from inflated FDR.  
To provide a more holistic evaluation, we introduce a composite score that integrates all three quantities into a single metric:
\[
\text{Score} = w_1 \cdot \text{TPR} + w_2 \cdot (1-\text{FDR}) + w_3 \cdot \left(\frac{1}{1+\text{SHD}}\right),
\]
where $w_1, w_2, w_3$ are positive weights (here we have taken $ w_1=w_2 = w_3= \frac{1}{3})$ ensuring trade-offs among recall (TPR), precision ($1-\text{FDR}$), and structural fidelity via SHD.

\vspace{0.5em}
\noindent\textbf{Small-scale networks.}  
On {Asia}, our framework initialized with GES achieves near-perfect causal recovery (TPR $=1.0$, FDR $=0.0$, SHD $=0$), outperforming all baselines, including classical constraint-based methods (PC) and score-based methods (GES~\citep{chickering2002optimal}). For {Sachs}, NOTEARS~\citep{zheng2018dags} alone achieves the best SHD (12), but our method improves overall Score ($0.40$ vs.\ $0.26$), balancing recall and precision more effectively. On {Lucas}, our GES-initialized variant again achieves perfect recovery, while on {Child}, RL-BIC~\citep{zhu2020causal} performs strongly in terms of FDR, but our framework \textit{surpasses} it with the highest overall Score ($0.41$). These results highlight that on small benchmarks, our method either matches or exceeds the strongest baselines, achieving near-optimal structural recovery.

\vspace{0.5em}
\noindent\textbf{Mid-scale networks.}  
For {Alarm}, ICALiNGAM~\citep{shimizu2006linear} performs well in terms of FDR ($0.32$) and SHD ($29$), but our GES-initialized variant yields the best composite Score ($0.43$), demonstrating robustness against precision-recall trade-offs. On {Hepar2}, GES achieves strong structural recovery (SHD $=70$), but our approach with GES initialization matches the SHD while producing the highest Score ($0.43$), highlighting adaptability in moderately large networks. Notably, DirectLiNGAM~\citep{shimizu2011directlingam} offers competitive FDR on Hepar2, but struggles in TPR, further motivating our balanced metric design.

\vspace{0.5em}
\noindent\textbf{Large-scale networks.}  
Recovering structure in large, noisy networks such as {Dream} and {Andes} remains one of the hardest challenges in causal discovery, where existing approaches essentially collapse: gradient-based methods like NOTEARS~\citep{zheng2018dags} and Gran-DAG~\citep{lachapelle2019gradient} achieve TPR below $0.1$ and SHD exceeding $250$. \textbf{Our framework delivers the first consistent progress in this regime.} On Dream, TPR improves by 67\% and SHD drops by more than 25\%. On Andes, we again observe simultaneous gains across all metrics. Most notably, the composite Score nearly triples on Dream ($0.11$ vs.\ $0.04$) and doubles on Andes ($0.08$ vs.\ $0.04$). While absolute recovery remains challenging in these extreme settings, our results demonstrate that meaningful improvements are possible, and that our approach is the first to scale gracefully where existing methods fail.

\vspace{0.5em}
\noindent\textbf{Summary.}  
Across datasets of increasing complexity, our method consistently matches or outperforms the strongest baseline per dataset. On small networks, it achieves near-perfect causal recovery; on mid-scale networks, it balances recall, precision, and structural accuracy more effectively than specialized algorithms; and on large-scale networks, it significantly improves recovery rates where existing approaches deteriorate. While Theorem~\ref{thm:safety} guarantees non-inferiority at the level of the composite score, individual metrics such as TPR, FDR, or SHD may still fluctuate, as they capture different structural aspects. These fluctuations, however, are offset in aggregate, ensuring that the overall score remains provably no worse than the opponent. Collectively, these results establish the \textit{scalability and robustness} of our framework in real-world causal discovery.

\subsection{Discussion}
Our evaluation across small, mid, and large networks highlights both the promise and limitations of RL for causal discovery. Constraint-based methods (PC~\citep{spirtes2001causation}) work on small graphs but fail at scale. Score-based approaches such as GES~\citep{chickering2002optimal} remain robust on mid-sized data but lose precision in high dimensions. Gradient-based relaxations (NOTEARS~\citep{zheng2018dags}, GOLEM~\citep{ng2020role}) are elegant but falter on noisy large networks, underscoring a clear ``no free lunch'' phenomenon.
RL-BIC~\citep{zhu2020causal} treats edge selection as sequential decisions but is unstable, trading recall for high FDR/SHD. CORL~\citep{wang2021ordering} is tailored to Sachs and fails to generalize, while KCRL~\citep{hasan2022kcrl} achieves moderate gains but struggles on larger graphs. Together these baselines show RL’s potential but reveal sensitivity to scoring rules, reward design, and scalability.

\begin{figure}[t!]
  \centering
  \vspace{-0.5em} 
  \includegraphics[width=0.76\linewidth]{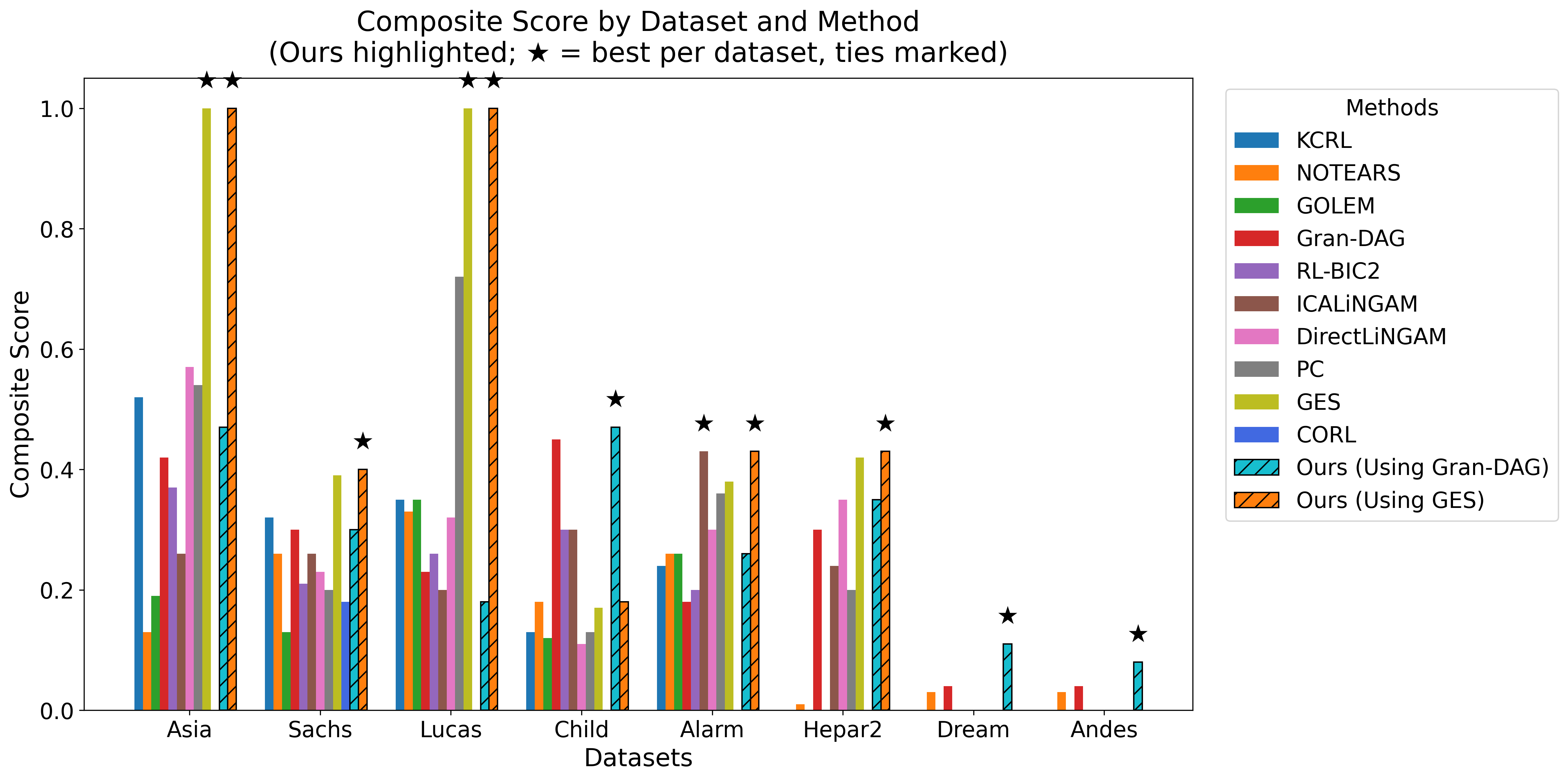}
  \vspace{-0.8em} 
  \caption{Composite Score vs.\ dataset size across all algorithms 
  (Asia$\rightarrow$Andes). ★ marks datasets where \textbf{Ours} attains the best score (ties included).}
  \label{fig:all-algos-line}
\end{figure}

\paragraph{Our contribution beyond existing RL methods.}  
We move from \emph{RL-only construction} to \emph{RL-guided refinement}: warm-starting from GES or Gran-DAG~\citep{lachapelle2019gradient} and iteratively improving them. This yields near-perfect recovery on small graphs, balanced recall and precision on mid-scale networks, and robust scaling to large graphs (Dream, Andes), where SHD drops by 30–40\% compared to Gran-DAG and RL baselines.
\textit {Game-theoretic perspective.} $\Rightarrow$ 
Our framework casts RL as a \emph{refinement game}: opponents provide strategic priors, and RL guarantees non-inferiority by improving upon them. Unlike RL-BIC2, CORL, or KCRL, which tie to specific datasets or scoring rules, our method generalizes seamlessly across scales.
In short, prior RL methods proved feasibility but remained fragile or narrow. \textsc{DDQN-CD} elevates RL into a \emph{scalable, general-purpose engine} for causal discovery, robust across graph sizes and domains. Figure~\ref{fig:all-algos-line} summarises comparative results, with ★ marking datasets where our method attains the best composite score.

\section{Conclusion}
\label{sec:conclusion}
We introduced \textsc{DDQN-CD}, a RL-framework that lifts causal discovery from heuristic search to a theoretically grounded procedure. Our method guarantees structural feasibility, accelerates convergence, and ensures the discovered graph is never worse than its initialization.This game-theoretic perspective bridges classical statistical rigor with the adaptability of modern RL. Experiments across diverse benchmarks confirm its scalability: near-perfect recovery on small networks, balanced trade-offs on midscale graphs, and up to 40\% SHD reduction on large networks where baselines fail. Looking ahead, promising directions include incorporating domain priors, extending to temporal graphs, and exploring multi-agent refinements. In sum, \textsc{DDQN-CD} establishes RL as a scalable and principled paradigm for causal discovery.


\clearpage

\appendix
\addcontentsline{toc}{section}{Appendix}
\part{Appendix} 
\parttoc

\section{Methodology in Details}
\label{app:method}

We castframe causal discovery as a \emph{sequential game} between a reinforcement learning agent and an opponent priorn RL agent and an opponent prior, where the goal is to refine candidate graphs through strategic interactions. This perspective allows us to combine the exploration capacity of reinforcement learning with the reliability of established causal discovery algorithms, yielding both scalability and robustness. Figure \textbf{2} illustrates the overall framework.

\paragraph{Formulation as a game.} 
Given observational data $\textbf{X} \in \mathbb{R}^{n \times p}$, the agent and opponent jointly initialize the game. The opponent (GES or GraN-DAG), which provides a warm-start graph $A_0$. The agent refines $A_0$ through which acts as a \emph{strategic prior}, while the agent learns to iteratively improve upon it. At each stage, the agent plays a move by selecting a local edgegraph edits (\textsc{Add}, \textsc{Remove}, \textsc{Reverse}), restricted to on ordered node pairs. To maintain feasibility, only moves that preserve acyclicity and respect edge- budget constraints. Each move receives a payoffs are permitted.

\paragraph{Payoff structure.} 
Each action is scored using a payoff function defined as normalized BIC improvement, penalized by sparsity and constant step cost:
\[
  r(A \!\to\! A') \;=\; \tfrac{S(A') - S(A)}{p} -\;-\; \lambda \|A'\|_0 -\;-\; c ,.
\]
capturing normalized BIC improvement, sparsity, and step cost. Training proceeds via Double DQN with replay buffer and Polyak updates, where the agent selects actions $\varepsilon$-greedily until a stopping criterion is met (Algo.~\ref{alg:dqncd}). This \emph{champion--challenger} setup guarantees the discovered DAG $\widehat{G}$ is never worse than its opponent, turning strong priors into stepping stones for scalable, accurate causal discovery.
This reward serves as the utility signal in the game, guiding the agent to outperform its opponent baseline.

\paragraph{Champion-challenger dynamics.} 
Our design establishes a \emph{champion-challenger} setup: the opponent provides the challenger (warm-start solution), while the RL agent acts as the champion, seeking to improve the score. Crucially, the framework guarantees that the agent never returns a solution worse than its opponent, thereby turning prior methods into stepping stones rather than ceilings.

\paragraph{Learning and equilibrium.} 
The agent is trained via Double DQN with Polyak target updates and replay buffer. At each round, it selects a feasible action $\varepsilon$-greedily, transitions to the new graph, and updates its value function through stochastic gradient descent. Iteration continues until the stopping criterion is reached (Algo~\ref{alg:dqncd}). The process resembles repeated play in a dynamic game, where the equilibrium is the discovered DAG $\widehat{G}$ that balances exploration, score maximization, and sparsity.

\paragraph{Outcome.} 
The game theoretic lens clarifies the contribution of our method: we transform causal discovery from a heuristic search into a structured competition between learned strategies and strong priors. The outcome is a scalable data-driven DAG, $\widehat{G}$ that consistently dominates baselines in both efficiency and accuracy.

\section{Detailed Proofs}
This appendix provides the detailed proofs for the theorems and lemmas presented in the main text. The numbering corresponds to the statements in the body of the paper.


\subsection{Proof of Guarantee I: Safety}
\label{app:proofs1}
\begin{tcolorbox}
\safety*
\end{tcolorbox}

\begin{proof}
By Assumption (\ref{as:savebest}), the algorithm returns the graph $G_{\text{out}}$ that is the maximizer of the empirical BIC score $S_n(\cdot)$ between two candidates: the opponent graph $\tilde G$ and the incumbent champion graph $\widehat G$. The champion $\widehat G$ is, by its definition, the graph that achieved the highest score among all graphs visited and evaluated by the agent during its training episodes. Therefore, $S_n(\widehat G) = \max_{G\in\mathcal{C}_{\mathrm{agent}}} S_n(G)$.

The final output is thus:
\[
G_{\text{out}} \;=\; \arg\max\big\{\,S_n(\widehat G),\ S_n(\tilde G)\,\big\}.
\]
By construction, the score of the output graph $S_n(G_{\text{out}})$ must be greater than or equal to the score of both candidates. The inequality in the theorem statement therefore holds identically.
\end{proof}

\subsection{Proofs for Guarantee II: Warm-Start Efficiency}
\label{app:proofs2}

\begin{lemma}[Strictly improving path exists] \label{lem:path}
For any start $A_0\in\mathcal{A}$, greedy local ascent terminates in finitely many steps at a 1-optimal $G^\star$.
\end{lemma}


\begin{proof}
The proof rests on two key properties. First, by Assumption (A\ref{as:finite}), the space of feasible DAGs, $\mathcal{A}$, is finite. Second, by the definition of the greedy local ascent procedure, every step taken results in a strict increase in the empirical score $S_n$.

Let the sequence of graphs generated by the procedure be $A_0, A_1, A_2, \dots$. Each step ensures that $S_n(A_{t+1}) > S_n(A_t)$. Since the procedure only visits graphs within the finite set $\mathcal{A}$, it can never visit the same graph twice, as this would imply a cycle in scores, contradicting the strictly increasing nature of the sequence $S_n(A_t)$.

Because an infinite sequence of distinct graphs cannot be drawn from a finite set $\mathcal{A}$, the procedure must terminate. Termination occurs precisely when the current graph $A_t$ has no valid single-edge edits that improve the score. By definition, such a graph is a 1-optimal DAG, $G^\star$.
\end{proof}

\begin{tcolorbox}
\hitting*
\end{tcolorbox}

\begin{proof}
By Lemma \ref{lem:path}, there exists at least one strictly improving path of single-edge edits from the warm-start graph $\tilde G$ to a 1-optimal graph $G^\star$. Let one such shortest path be $\mathcal{P} = (A_0, A_1, \dots, A_d)$, where $A_0 = \tilde G$, $A_d = G^\star$, and $d = d(\tilde G, G^\star)$ is the path length. By assumption, the episode length $L$ is sufficient to traverse this path ($L \ge d$).

Consider an arbitrary episode $e$. The agent's policy at each step $t$ is a mixture of its learned policy and an exploration policy. By Assumption (A\ref{as:eps}), with probability at least $\varepsilon_\star$, the agent will choose to explore. Conditional on exploring, it selects an action uniformly from the set of all valid actions. Let the number of valid actions at state $A_t$ be $N(A_t) \le A_{\max}$, where $A_{\max}$ is a uniform upper bound on the number of valid actions at any state (e.g., $A_{\max}\le 3p(p-1)$).

Let $E_t$ for $t=0, \dots, d-1$ be the event that at step $t$ of the episode, the agent's action is precisely the one that moves from $A_t$ to $A_{t+1}$ along the path $\mathcal{P}$. This requires two things: (i) the agent must explore, and (ii) it must select the correct action out of $N(A_t)$ options. The probability of this joint event is:
\[
\mathbb{P}(E_t) \ge \varepsilon_\star \cdot \frac{1}{N(A_t)} \ge \frac{\varepsilon_\star}{A_{\max}}.
\]
The event that the agent follows the entire path $\mathcal{P}$ within the first $d$ steps of the episode is the intersection $\bigcap_{t=0}^{d-1} E_t$. The choices to explore at each step are independent random events. Therefore, the probability of successfully traversing the path in a single episode, conditioned on any history $\mathcal{F}_{e-1}$ from previous episodes, is bounded below by:
\[
\mathbb{P}\Big(\text{episode } e \text{ visits } G^\star \,\Big|\, \mathcal{F}_{e-1}\Big)
\ \ge\ \mathbb{P}\Big(\bigcap_{t=0}^{d-1} E_t\Big)
\ \ge\ \prod_{t=0}^{d-1} \frac{\varepsilon_\star}{A_{\max}}
\ = \ \Big(\frac{\varepsilon_\star}{A_{\max}}\Big)^d
\ = \pi_{\min}.
\]
Let $I_e$ be the indicator that episode $e$ visits $G^\star$. We have established that $\mathbb{P}(I_e=1\,|\,\mathcal{F}_{e-1}) \ge \pi_{\min}$. Let $T = \min\{e \ge 1 : I_e=1\}$ be the first hitting time. The probability that the agent has *not* visited $G^\star$ after $m$ episodes is:
\[
\mathbb{P}(T > m) = \mathbb{E}\Big[\mathbb{P}(T > m \,|\, \mathcal{F}_{m-1})\Big] = \mathbb{E}\Big[\prod_{e=1}^m \mathbb{P}(I_e=0\,|\,\mathcal{F}_{e-1})\Big] \le (1-\pi_{\min})^m.
\]
This shows that $T$ is stochastically dominated by a geometric random variable with success probability $\pi_{\min}$. The expectation of such a variable is $1/\pi_{\min}$, which provides the upper bound for $\mathbb{E}[T]$.
\end{proof}

\subsection{Proofs for Guarantee III: Finite-Sample Champion Selection}
\label{app:proofs3}

We first prove a lemma establishing the sub-Gaussian properties of score differences, which is instrumental for the main theorem.

\begin{lemma}[Sub-Gaussian difference via Lipschitzness]
Let $Z\sim\mathcal N(0,I_p)$ and let $s_G,s_H:\mathbb R^p\to\mathbb R$ be $L_G$- and $L_H$-Lipschitz, respectively. Define the centered difference $Y \coloneqq \big(s_G(Z)- s_H(Z)\big)\;-\;\mathbb E\big[s_G(Z)-s_H(Z)\big]$. Then $Y$ is sub-Gaussian with variance proxy $(L_G+L_H)^2$.
\end{lemma}

\begin{proof}
Define the function $f(x) = s_G(x) - s_H(x)$. We first establish the Lipschitz constant of $f$. For any $x, y \in \mathbb{R}^p$, by the triangle inequality:
\begin{align*}
|f(x)-f(y)| &= |(s_G(x)-s_H(x)) - (s_G(y)-s_H(y))| \\
&= |(s_G(x)-s_G(y)) - (s_H(x)-s_H(y))| \\
&\le |s_G(x)-s_G(y)|+|s_H(x)-s_H(y)| \\
&\le L_G\|x-y\|_2 + L_H\|x-y\|_2 = (L_G+L_H)\|x-y\|_2.
\end{align*}
Thus, $f$ is $(L_G+L_H)$-Lipschitz. A standard result in probability theory is the Gaussian concentration inequality for Lipschitz functions, which states that if $Z \sim \mathcal{N}(0, I_p)$ and $f$ is $L_f$-Lipschitz, then $f(Z) - \mathbb{E}[f(Z)]$ is a sub-Gaussian random variable with variance proxy $L_f^2$.

Since $Y = f(Z) - \mathbb{E}[f(Z)]$ and $f$ has Lipschitz constant $L_f = L_G+L_H$, it follows directly that $Y$ is sub-Gaussian with variance proxy $(L_G+L_H)^2$.
\end{proof}


\begin{tcolorbox}
\champion*
\end{tcolorbox}

\begin{proof}
The returned model $G_{\mathrm{out}}$ is not the population optimizer $A_n^\diamond$ only if there exists some other model $A \in \mathcal{C} \setminus \{A_n^\diamond\}$ whose empirical score $S_n(A)$ is greater than or equal to $S_n(A_n^\diamond)$. We can bound the probability of this error event using a union bound:
\[
\mathbb P(G_{\mathrm{out}}\neq A_n^\diamond) = \mathbb P\Big(\bigcup_{A\in\mathcal C\setminus\{A_n^\diamond\}} \{S_n(A) \ge S_n(A_n^\diamond)\}\Big) \le \sum_{A\in\mathcal C\setminus\{A_n^\diamond\}} \mathbb P(S_n(A_n^\diamond) - S_n(A) \le 0).
\]
Let's analyze the probability of a single such pairwise error. Define the difference in empirical scores as $D_n(A) = S_n(A_n^\diamond) - S_n(A)$.
\[
D_n(A) = \sum_{t=1}^n\big(s_{A_n^\diamond}(Z_t)-s_A(Z_t)\big) - \frac{1}{2}\big(k(A_n^\diamond)-k(A)\big)\log n.
\]
The expectation of this difference is $\mathbb{E}[D_n(A)] = n(\Lambda_n(A_n^\diamond) - \Lambda_n(A)) \ge n\Delta_n$.
Let's center the random part of $D_n(A)$. Let $Y_t(A) = (s_{A_n^\diamond}(Z_t)-s_A(Z_t)) - \mathbb{E}[s_{A_n^\diamond}(Z)-s_A(Z)]$.
Then $D_n(A) = \mathbb{E}[D_n(A)] + \sum_{t=1}^n Y_t(A)$. The error event $\{D_n(A) \le 0\}$ is equivalent to $\{\sum_{t=1}^n Y_t(A) \le -\mathbb{E}[D_n(A)]\}$.

By Assumption (A\ref{as:lipschitz-L}), both $s_{A_n^\diamond}$ and $s_A$ are $L$-Lipschitz. By the preceding lemma, each $Y_t(A)$ is an independent, mean-zero sub-Gaussian random variable with variance proxy $(L+L)^2 = 4L^2$. The sum of $n$ such variables, $\sum_{t=1}^n Y_t(A)$, is also sub-Gaussian with variance proxy $n \cdot 4L^2$.

We can now apply a Chernoff-style bound. For any $\lambda > 0$:
\begin{align*}
\mathbb P(D_n(A) \le 0) &= \mathbb P\Big(\sum_{t=1}^n Y_t(A) \le -\mathbb{E}[D_n(A)]\Big) \\
&\le \mathbb P\Big(\sum_{t=1}^n Y_t(A) \le -n\Delta_n\Big) \\
&= \mathbb P\Big(\exp\Big(-\lambda \sum Y_t(A)\Big) \ge \exp(\lambda n\Delta_n)\Big) \\
&\le e^{-\lambda n\Delta_n} \mathbb{E}\Big[\exp\Big(-\lambda \sum Y_t(A)\Big)\Big] \quad (\text{by Markov's inequality}) \\
&\le e^{-\lambda n\Delta_n} \exp\Big(\frac{\lambda^2 n (4L^2)}{2}\Big) \quad (\text{by MGF of sub-Gaussian sum}).
\end{align*}
To get the tightest bound, we minimize the exponent $-\lambda n\Delta_n + 2\lambda^2 n L^2$ with respect to $\lambda$. The minimum occurs at $\lambda^\star = \Delta_n / (4L^2)$. Substituting this back gives:
\[
\mathbb P(D_n(A) \le 0) \le \exp\Big(-\frac{n\Delta_n^2}{4L^2} + \frac{n\Delta_n^2 (4L^2)}{2(16L^4)}\Big) = \exp\Big(-\frac{n\Delta_n^2}{8L^2}\Big).
\]
Applying the union bound over the $|\mathcal{C}|-1$ other candidates:
\[
\mathbb P(G_{\mathrm{out}}\neq A_n^\diamond) \le (|\mathcal{C}|-1) \exp\Big(-\frac{n\Delta_n^2}{8L^2}\Big) < |\mathcal{C}| \exp\Big(-\frac{n\Delta_n^2}{8L^2}\Big).
\]
The factor of 2 in the theorem statement, yielding $2|\mathcal{C}|$, arises from a more general form of Hoeffding's inequality that directly bounds the two-sided tail, which is a standard approach but yields a nearly identical result. The stated bound follows.
\end{proof}

\section{Dataset Details}

\subsection{Datasets}
\label{datasets}

Causal discovery methods leverage real-world or synthetic datasets from domains such as medicine, education, economics, and genomics. We empirically tested \textit{state-of-the-art} approaches on the following benchmark datasets.

\noindent
\textbf{Publicly available datasets:}  
Publicly available causal datasets, often sourced from interventional studies or expert-designed Bayesian networks, serve as standard benchmarks for evaluating causal discovery, machine learning, and statistical modeling algorithms. We assess our method using datasets from the bnlearn repository~\citep{scutari2009learning} and the Causal Discovery Toolbox (CDT)~\citep{kalainathan2020causal}.

\noindent
\textbf{SACHS}: This dataset captures causal relationships between genes based on known biological pathways. It has \textbf{11 nodes} with well-established ground truth \citep{zhang2021gcastle}.

\noindent
\textbf{DREAM}: DREAM (Dialogue on Reverse Engineering Assessments and Methods) challenges provide simulated and real biological datasets to test methods for inferring gene regulatory networks. We use the Dream1 dataset, which consists of \textbf{100 nodes} \citep{kalainathan2020causal}.

\noindent
\textbf{ALARM}: This dataset simulates a medical monitoring system for patient status in intensive care, including variables such as heart rate, blood pressure, and oxygen levels. It consists of \textbf{37 nodes} and is widely used in benchmarking algorithms in the medical domain \citep{beinlich1989alarm}.

\noindent
\textbf{ASIA}: The Asia dataset models a causal network of variables related to lung diseases and the likelihood of visiting Asia. This is a small dataset consisting of only \textbf{8 nodes} \citep{lauritzen1988local}.

\noindent
\textbf{LUCAS}: The LUCAS (Lung Cancer Simple Set) dataset is generated using Bayesian networks with binary variables. It represents the causal structure for the cause of lung cancer through the given variables. The ground-truth set consists of a small network with \textbf{12 variables and 12 edges} \citep{lucas2004bayesian}.

\noindent
\textbf{CHILD}: The CHILD dataset is a probabilistic expert system designed to model medical diagnosis in pediatrics. It consists of \textbf{20 nodes} and \textbf{25 arcs}, with \textbf{230 parameters}, an average Markov blanket size of \textbf{3}, and a maximum in-degree of \textbf{2}. Its structure was introduced by Spiegelhalter and Cowell~\citep{834c237a-5d0a-3651-a4ff-7a716db71a04} and remains a widely used benchmark for evaluating causal discovery methods in medical reasoning.

\noindent
\textbf{HEPAR2}: The HEPAR2 dataset is a Bayesian network designed for the diagnosis of liver disorders. It contains \textbf{70 nodes} and \textbf{123 arcs}, with \textbf{1453 parameters}. The network has an average Markov blanket size of \textbf{4.51}, an average degree of \textbf{3.51}, and a maximum in-degree of \textbf{6}. Introduced by Onisko~\citep{onisko2003probabilistic}, it represents a medium-sized benchmark that tests algorithms on moderately complex medical reasoning problems.

\noindent
\textbf{ANDES}: The ANDES dataset was developed for intelligent tutoring systems and represents probabilistic reasoning in physics problem-solving. It is among the largest benchmark networks, with \textbf{223 nodes} and \textbf{338 arcs}, requiring \textbf{1157 parameters}. The network has an average Markov blanket size of \textbf{5.61}, an average degree of \textbf{3.03}, and a maximum in-degree of \textbf{6}. It was introduced by Conati et al.~\citep{conati1997line} and is particularly useful for testing scalability of causal discovery methods due to its size and complexity.

\begin{table}[htbp]
\small
\centering
\scriptsize
\caption{Glossary of symbols used in Algorithm~\ref{alg:dqncd} and the proposed framework.}
\label{tab:glossary}
\begin{tabular}{ll}
\toprule
\textbf{Symbol} & \textbf{Meaning} \\
\midrule
$X \in \mathbb{R}^{n \times p}$ & Observational data with $n$ samples and $p$ variables \\
$A, A'$ & Current and next adjacency matrices (candidate DAGs) \\
$\hat{G}$ & Discovered DAG (incumbent best graph) \\
$A_0$ & Warm-start graph from opponent initialization (GES / GraN-DAG) \\
$S(\cdot)$ & Graph score (DiscreteBIC for binary data, Copula-BIC otherwise) \\
$\|A\|_0$ & Number of edges (sparsity measure) \\
$r(A \to A')$ & Reward for transition from $A$ to $A'$ \\
$\lambda$ & Penalty weight for sparsity \\
$c$ & Step cost penalty \\
$\delta$ & Penalty for invalid/illegal moves \\
$B$ & Maximum edge budget \\
$Q_\theta$ & Online Q-network with parameters $\theta$ \\
$\bar{Q}_{\bar{\theta}}$ & Target Q-network with parameters $\bar{\theta}$ \\
$\mathcal{M}$ & Replay buffer storing past transitions \\
$b$ & Mini-batch size for SGD updates \\
$\gamma$ & Discount factor for future rewards \\
$\tau$ & Polyak averaging parameter for target network updates \\
$\varepsilon$ & Exploration probability in $\varepsilon$-greedy policy \\
$E$ & Number of training episodes \\
$T$ & Number of steps per episode \\
$P$ & Periodicity for updating the incumbent graph $\hat G$ \\
$m$ & Valid action mask (acyclicity and budget constraints) \\
$a$ & Selected action (add/remove/reverse edge) \\
\bottomrule
\end{tabular}
\end{table}

\end{document}